# An Approach for Efficient Neural Architecture Search Space Definition

Léo Pouy[1], Fouad Khenfri[1], Patrick Leserf[1], Chokri Mraidha[2], Cherif Larouci[1]

Ecole Supérieure des Techniques Aéronautiques et de Construction Automobile[1,],
Université Paris-Saclay, CEA, List, F-91120, Palaiseau, France [2]
leo.pouy@estaca.fr, fouad.khenfri@estaca.fr, patrick.leserf@estaca.fr, chokri.mraidha@cea.fr, cherif.larouci@estaca.fr

**Abstract**

As we advance in the fast-growing era of Machine Learning, various new and more complex neural architectures are arising to tackle problem more efficiently. On the one hand their efficient usage requires advanced knowledge and expertise, which is most of the time difficult to find on the labor market. On the other hand, searching for an optimized neural architecture is a time-consuming task when it is performed manually using a trial and error approach. Hence, a method and a tool support is needed to assist users of neural architectures, leading to an eagerness in the field of Automatic Machine Learning (AutoML). When it comes to Deep Learning, an important part of AutoML is the Neural Architecture Search (NAS). In this paper, we propose a novel cell-based hierarchical search space, easy to comprehend and manipulate. The objectives of the proposed approach are to optimize the search-time and to be general enough to handle most of state of the art Convolutional Neural Networks (CNN) architectures.

# Introduction

Machine Learning (ML), especially Deep Learning (DL), is a field that has seen a high growth of interest in the last decades, in research as well as in the industry. However, as Deep Learning algorithms are black boxes, the development of a deep learning model is very empirical and a lot of experience is needed to find the right Neural Architecture and fine tune its HyperParameters. That is where Automatic Machine Learning (AutoML) comes in. Indeed, AutoML aims at automating the tedious tasks when developing a ML model, so less human input and experience is needed to build efficient models. This is a promising field that has seen a lot of innovation in the last years (Xin et. Al. 2021).

For DL, the AutoML optimization is separated into two problems: HyperParameter Optimization (HPO) and Neural Architecture Search (NAS), the first being related to training parameters (such as the learning rate, the momentum, the batch size, etc.), while the second is related to the model parameters (such as the number of layers, the number of neurons for a densely connected layer, the kernel size of the convolution layers, etc.).

The NAS when designing the DL model, not only have an influence on the accuracy of the model but also defines the execution time and the overall computational demands, in training as well as once in deployment. Therefore, it is necessary for a number of real world constrained problem, such as embedded systems, to take the target's resource constraints into account while performing NAS, so the selected model meets the possible restrictions (maximum execution time, limited storage, limited random-access memory, etc.).

To perform a NAS, two elements are needed. First a search space to explore, containing all the possible architectures that can be found. Then an optimization algorithm is used to explore this search space to find the best or a satisfactory solution. However, search spaces for NAS can rapidly become extremely large (for example, the search space proposed in Pham et. al. has over $10^{15}$ candidate architectures), and take long time to converge to a solution (Zoph et.al. (2018) takes 2000 GPU days). To address this problem, one can either develop more efficient optimization algorithm to explore the search space, or reduce the search space. The latter implies needs to be carefully done, as reducing the search space might lead to excluding optimal solutions. Another downside is that limiting the search space implies limiting the freedom of a user to orient the search toward certain architectures.

Therefore, this paper focuses on the definition of an adaptive base search space that is general enough to find state-of-the-art architecture.

The rest of the paper is organized as follows:
- A background and related work.
- A casual description of the search space structure.
- A formalization of the search space and the way to generate architectures.
- A description of the optimization loop intended and the one used in preliminary results.
- A preliminary implementation and results obtained
- Conclusion and future work.

# Background

To define good properties for an efficient search space, this section provides an overlook of the history of well performing architectures. A focus has been made on Convolutional Neural Network (CNN) architectures which are the most known DL models for computer vision.

The basis of CNN was introduced by LeCun et. al. (1989) for handwritten digits and zip code recognition with a first, simple, architecture of basic convolutions put one after the other. Demonstrating that convolution can greatly enhance the accuracy of a Neural Network, so our search space has convolution as it basic element.

Then came AlexNet (Krizhevsky et. al. 2017) in 2012 demonstrated that having a deeper model, meaning the architecture has more layers, generally increases the performance of the model. Moreover, they used GPU to compensate for the increase in computational cost. Hence the ability to easily control the number of layers in the search space.

Another historic architecture is the VGG model (Simonyan et. al. 2014), which added the notion of reducing the resolution of the images when going deeper in the layers. This enabled to extract finer features from the image. So the includes capability to reduce the spatial resolution.

In 2014, Szegedy et. al. proposed GoogLeNet, a model based on Inception modules. Those modules perform convolution in parallel, effectively making the model wider rather than deeper. This enable to reduce the overfitting. This motivated the choice to design the search space such that convolutions can be done in parallel. We call this parallelization "pipelines".

Finally came ResNet (He et. al. 2015) that introduced the notion of "skipping" layers. A skip is the operation of adding (or concatenating) the result of a previous layer to the result of the current layer, retaining the features extracted before throughout the model. So we implemented that some of the pipelines can be identity operation, recreating the ResNet feature.

Lately, there have been gain in interest to go to simpler models (Hasanpour et. al. 2018 and Assiri 2020) that still performs state-of-the-art accuracy. These works benefit from a high amount of regularization enabling them to train a lot without overfitting. This is why the option to put regularization component in the search space has been added.

## Related work

This section describes the three main categories of search space for NAS.

Chen et al. (2016), inspired by transfer learning (Yosinski et. al. 2014), proposed an architecture generation based on altering existing networks, making them deeper and/or wider at each iteration. This has the effect to retain the knowledge of the previous model while improving its capabilities. However, this method is not suited for our problem as it needs an existing model. Indeed, we aim at developing a search space that can include architecture that are not derived from human-crafted architecture.

Following the observation that the repetition of modules (Szegedy et. al. 2014) performs really well, Pham et. al. (2018), Zhong et.al (2018) and Zoph et. al. (2018) developped a cell-based search space relying on a succession of repeating cells containing the operations. With this type of search space, only the components of the cell are searched, then the model is crafted by repeating the cell. Therefore, it is possible to perform NAS on very deep models without increasing the complexity of the search space. This is why we chose to develop a cell-based search space, as it gives a good control over the complexity of the search space.

Hierarchical search space (Liu H. et. al. 2019, Liu C. et. al. 2019, and Zhang et. al. 2021) further extend the cell-based search space. It came from the observation that controlling the resolution of the images throughout the network can highly increase the performance (Long et. al. 2015 and Ronneberg et. al. 2015). Therefore, a hierarchical search space is a cell-based search space with the notion of spatial resolution control. This means that throughout the model, the spatial resolution can stay the same, or be downsampled (halving the resolution by 2 and doubling the number of filters), or upsampled (doubling the resolution and halving the number of filters). We chose to tackle this issue in our search space as well, as these NAS frameworks define the current state of the art in term of Computer Vision AutoML.

Therefore, this work has a similar search space to the one used by Liu C. et al. (2019) and Zhang et. al. (2021), but tries to make it more flexible to enhance the scope of obtainable architectures while keeping the complexity of the search space mastered.

## Description of the search space

This proposed search space focuses on the feature extraction part of the deep learning model. Consequently, if the feature extraction is not sufficient by itself, an output module can be set manually, for example a fully connected layer (FC).

In this section, the search space will be presented beginning by its highest level, then the structure of a cell is shown, and finally the fundamental component, the blocks, will be described.

### A cell based search space

Figure 1 presents the overall structure of the cell-based search space. It is constituted of $L_c$ layers of cells in series. In each cell layer, one cell is selected from the $N_c$ possible cells. Each cell can be configured with $P_c$ options as "a normal cell" which keeps the spatial resolution of the input tensor, also "a reduction cell" which divides the spatial resolution by 2 and multiplies the number of filters by 2. Thus, we have to evaluate $(P_c \times N_c)^{L_c}$ architectures.

Before selecting the cells, they must first be constructed. To do so, the cell is composed of two parts: a convolutional part (making the operations) and a reduction part (controlling the spatial resolution).

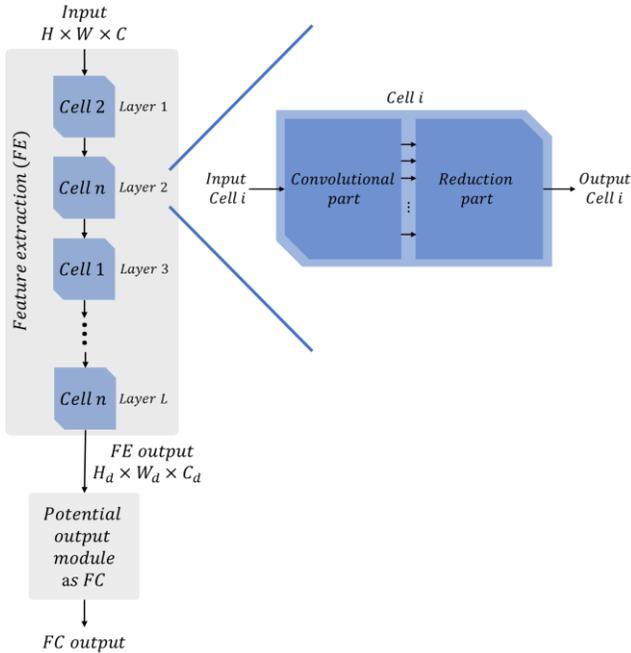

Figure 1 Structure of the search space

### Inside of a cell

The convolutional part, shown in Figure 2, holds the operations. It is composed of $L_P$ parallel pipelines. Each pipeline contains up to $L_B$ layers of blocks in series. In each block layer, one block is selected from the $N_B$ possible blocks. As a block have $P_B$ options that can be selected as well, the number of possible pipeline architectures is $(P_B \times N_B)^{L_B}$. Thus, the number of possible convolutional part architectures is $(P_B \times N_B)^{L_B * \tilde L_B}$.

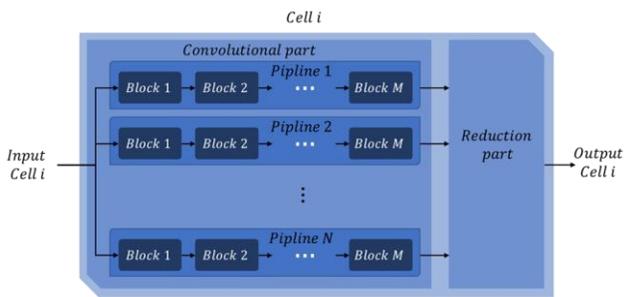

Figure 2 Convolution part

The reduction part presented in Figure 3 controls the spatial resolution of the input tensor using three methods downsampling, upsampling or samesampling:
- The downsampling halves the dimensions of the tensors (height and length of the image) and double the number of filters.
- The upsampling, as opposed to the downsampling, doubles the dimensions and halves the number of filters.
- Finally, the samesampling does not affect the dimensions of the tensors or the number of filters (i.e., reduction part with only merge operation).

The three methods mentioned above are selected at cell level. However, the type of block and its position relative to the merge operation are defined in the reduction part. Thus, the number of possible reduction part architectures is $P_r \times N_r^{L_r} + P_r \times N_r$. The first term presents the possibility to create $L_r$ layers of parallel reduction block before the merge operation (see Figure 3.a). The second term presents the possibility to create one reduction block layer after the merge operation (Figure 3.b). In each layer, one reduction block is selected from the $N_r$ blocks. The $P_r$ options define the type of the merge operations as addition or concatenate.

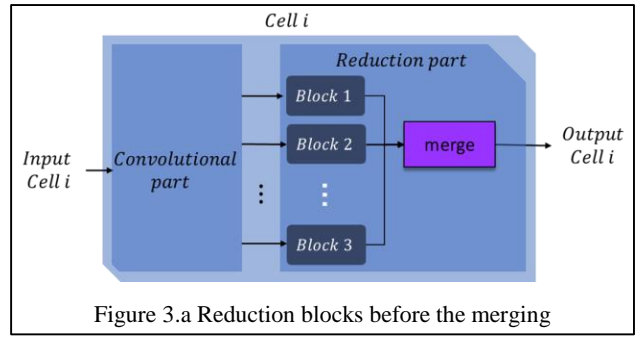

Figure 3.a Reduction blocks before the merging

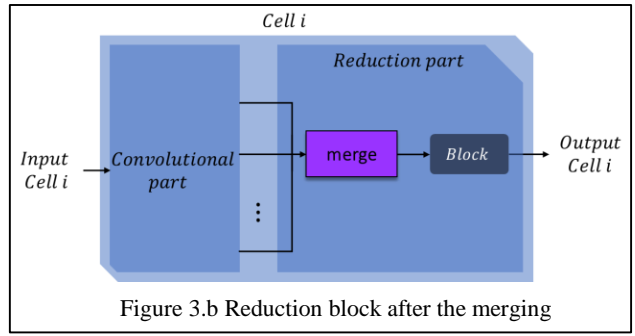

Figure 3.b Reduction block after the merging

Figure 3 Reduction part

### Blocks

A block is a basic element in our architecture and is defined by the user. A block can contain a single operation or a group of operations. An operation can be a convolution, polling, dense, or merge. A block can be assigned some options to create different configurations of the blocks. Therefore, the user needs to give a set of blocks that can be selected in the pipelines of the convolution part as well as a set of reduction blocks for the reduction part.

The Figure 4 shows three examples of blocks. The top one holds a single 3x3 convolution as an operation (represented by a square on the figure) while the middle one holds a 5x5 convolution followed by options (represented by a circle), which are a batch-normalization and a ReLU activation. Whereas the bottom one contains a 3x3 convolution and a 3x3 Max-Pooling in parallel that are concatenated together before passing an optional ReLU activation.

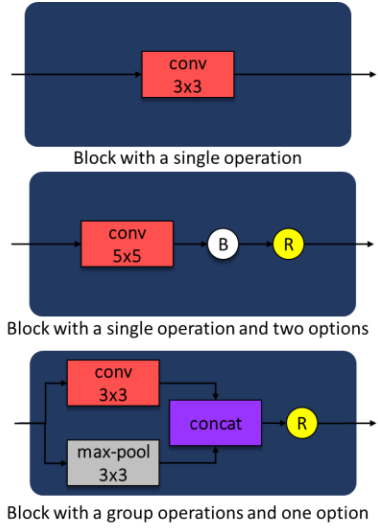

Figure 4 Example of blocks

## Formalization of the search space

This section explains how the search space is mathematically explored, and how the architecture is generated from it.

In our approach, the AutoML problem is formalized as a constrained multi-objective optimization problem described by the following equation:

$$\min_{X_{min} \leq X \leq X_{max}} F(X) = \begin{cases} f_1(X) \\ f_2(X) \\ \vdots \\ f_m(X) \end{cases}$$

$$\text{subject to}$$
$$h_i(X) = 0, i = 1,2,\ldots,n_h$$
$$g_i(X) \leq 0, i = 1,2,\ldots,n_g$$

Where $F(X)$ is the set of the objective functions that needs to be minimized. The functions $h_i(X)$ and $g_i(X)$ are respectively the constraints of equality and inequalities, and give further boundaries to the search space.

The vector $X$ represents the architecture of the feature extraction and is defined such as:
$X = [x, x_{11}, \ldots, x_{1L_{P_1}}, x_{21}, \ldots, x_{2L_{P_2}+1}, \ldots, x_{N_c1}, \ldots, x_{N_cL_{Pn}+1}]$
Where:
- $0 \leq x < (P_c \times N_c)^{L_c}$ represents the allocation of $P_c \times N_c$ cells into $L_c$ layers (see Figure 1)
- $x_{ij}$ is the parameter of $i^{th}$ cell where $j = 1, \ldots, L_{Pi}$ presents the pipeline of convolution part $0 \leq x_{ij} < (P_B \times N_B)^{L_B}$ and $j = L_{Pi} + 1$ is the reduction part $0 \leq x_{ij} < P_r \times N_r^{L_r} + P_r \times N_r$.

## Optimization loop

To test the search space, multi-objective Evolutionary Algorithm (EA) has been used. The search space is meant to run with two loops, as shown in Figure 5: a first loop creates $N_C$ cells, while the inner loop puts the cells in the layers and selects the $P_C$ options.

This double optimization loop enables to separate the search space in, on the one hand, the search of $N_C$ cells with a complexity of $(P_B \times N_B)^{L_B} + P_r \times N_r \times (N_r^{L_r} + 1)$ each and, on the other hand, the search of the structure with a complexity of $(P_C * N_C)^{L_C}$.

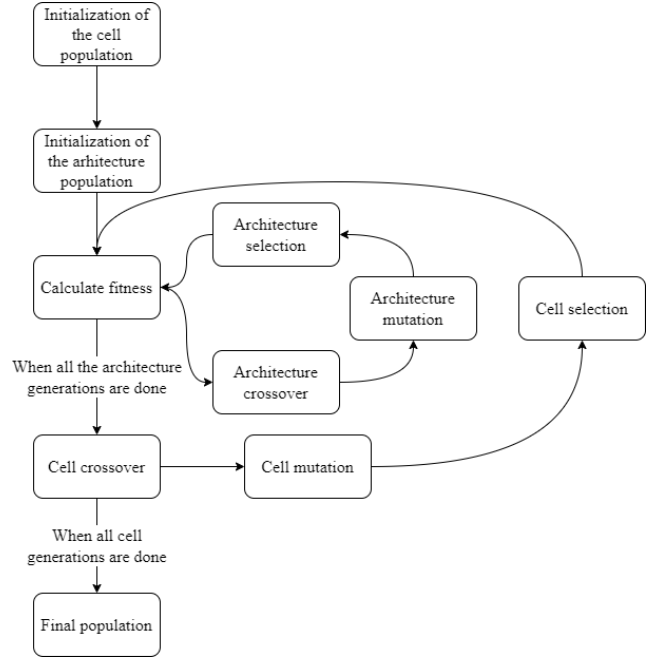

Figure 5 Optimization loop

However, for preliminary testing, the search has been made in a single loop where the EA create populations and optimize directly on the vector X.

## Preliminary testing implementation

We developed the implementation of the search space using Keras (Chollet F. et al. 2015), running on a NVIDIA Quadro RTX 6000. To perform the multi-objective EA, the pymoo (Blank and Deb 2020) library has been used.

**Search space parameters**

To define the search space, the blocks available must first be chosen. All the blocks used, illustrated in Figure 6, have the same structure. They differ by the operation they hold and the options they can have.

The operations available, commonly used in CNNs, are the following (all these operations are made with padding):
- 2D Convolution with a 3x3 kernel
- 2D Convolution with a 5x5 kernel
- 2D Convolution with a 7x7 kernel
- 2D MaxPooling with a 3x3 pool size
- Depthwise separable 2D Convolution with a 7x7 kernel.

Concerning the options, they concern the presence, or not, of a batch normalization layer (B) or a ReLU activation layer (R). There is also an option to skip the block, meaning it becomes an identity block. Therefore, the options can be as following:
- Skip
- No batch normalization & no activation function
- Batch normalization & no activation function
- Batch normalization & activation function before the operation
- Batch normalization & activation function after the operation
- No batch normalization & activation function before the operation
- No batch normalization & activation function after the operation.

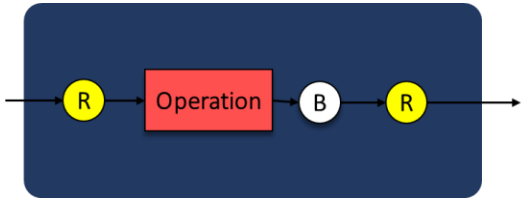

Figure 6 Block in the implementation

When it comes to the reduction blocks, they have no options and can either be:
- a 2D MaxPooling with a 3x3 pool size and a stride of 2x2
- a 2D Convolution with a 1x1 kernel and a stride of 2x2
- an identity operation (skip).

So in this case, they can only perform downsampling or samesampling.

For the structure, the parameters are as follows:
- There are 4 blocks in a pipeline,
- There are 3 pipelines in the convolution part,
- There are 2 cells,
- Two densely connected layer, with a dropout layer in-between, are put at the end of the model as a classifier. The first has 2048 neurons with a ReLU activation function while the other has 10 neurons with Softmax activation.

**HyperParameters**

The HPs when fitting the model are as follows:
- The optimizer is the default Adam algorithm of Keras,
- The batch size is 128,
- The model is trained for 10 epochs,
- The dropout layer has a 0.7 ratio.

**Objective functions**

There are two objectives, the error on the dataset at the end of the training and the number of parameters, such as:
- $f_1(X) = 1 - ACC(X)$, where $ACC(X)$ is the categorical accuracy of a given model $X$,
- $f_2(X) = \text{Param}(X)/\text{TotalParam}$, where $Param(X)$ is the number of parameters of a given model $X$ and is the maximum number of parameters allowed (it is set to 150.000.000).

There is also a constraint of inequality $g_2(X) = f_2(X) - 1 \leq 0$, so the models selected cannot have a higher number of parameters than $TotalParam$.

**Evolutionary algorithm parameters**

The parameters for the EA are:
- A population size of 20
- 1000 generations
- The crossover probability is of 1 and the eta is 3
- The mutation has a probability of 1 and an eta of 3.

## Results on MNIST

The previously presented implementation has been run twice on MNIST (LeCun et. al. 2010), and obtained the Pareto front given in Figure 7. The NAS found architectures with similar performance in both runs, showing the steadiness of the method.

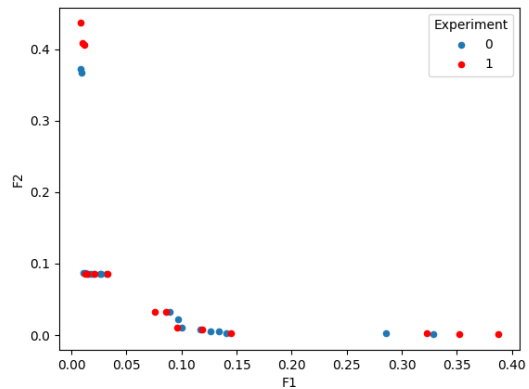

Figure 7 Pareto front for MNIST experiment

Then the model the furthest to the bottom-right of the Pareto front, which has 15 million trainable parameters, was trained for an additional 100 epochs and achieved an accuracy of 99,35% / error of 0,65%, without fine-tuning the hyperparamters.

Therefore, while the method did not reach the accuracy of recent state-of-the art human-crafted architecture, as it can be seen in Table 1, obtained results are still good enough to validate the viability of the search space, and are promising to pursue research in this direction.

| Test Error | Model |
|---|---|
| 0.16% | Efficient-CapsNet (Mazzia et. al. 2021) |
| 0.17% | SOPCNN (Assiri et. al. 2020) |
| 0,25% | SimpleNet (Hassanpour et. al. 2018) |
| 0.50% | ReNet (Visin et.al. 2015) |
| 0.60% | Convolutional Tsetlin Machine (Granmo et.al. 2019) |
| 0.70% | Deep Fried Convnets (Yang et.al. 2015) |
| **0.35%** | **Our method** |

Table 1. MNIST state-of-the-art

## Conclusion and future work

In this paper, we presented a method for the definition of adaptable search spaces. Obtained search spaces can be small and sufficiently general to include most state of the art architectures. Furthermore, it can also be enhanced in size (at the expense of search time/convergence difficulty) and it is possible to implement specific blocks that could help improve the performance, such as SENets (Hu et. al .2020).

We implemented a preliminary version of our NAS framework and validated that it is indeed working. For future work, we intend to improve further the search space, and more importantly the searching strategy to make the framework achieve close to the state of the art accuracy on MNIST as well as on other datasets, such as CIFAR10, CIFAR100, ImageNet, Cityscapes.